\documentclass[lettersize,journal]{IEEEtran}
\usepackage[numbers]{natbib}
\usepackage{lipsum} %
\usepackage{amsmath,amsfonts}
\usepackage{algorithmic}
\usepackage{algorithm}
\usepackage{array}
\usepackage[caption=false,font=normalsize,labelfont=sf,textfont=sf]{subfig}
\usepackage{textcomp}
\usepackage{stfloats}
\usepackage{url}
\usepackage{verbatim}
\usepackage{graphicx}
\usepackage{comment}
\usepackage{multirow}
\usepackage{xcolor}
\usepackage{amssymb}
\usepackage{booktabs}
\usepackage[hidelinks]{hyperref}\usepackage{etoolbox,xspace}
\usepackage{cuted}
\usepackage{capt-of}
\usepackage{enumitem}

\newcommand{\methodName}{DiffGaze\xspace}

\usepackage{graphicx}
\usepackage{etoolbox}

\newcommand{\insertfig}{
\includegraphics[width=\textwidth]{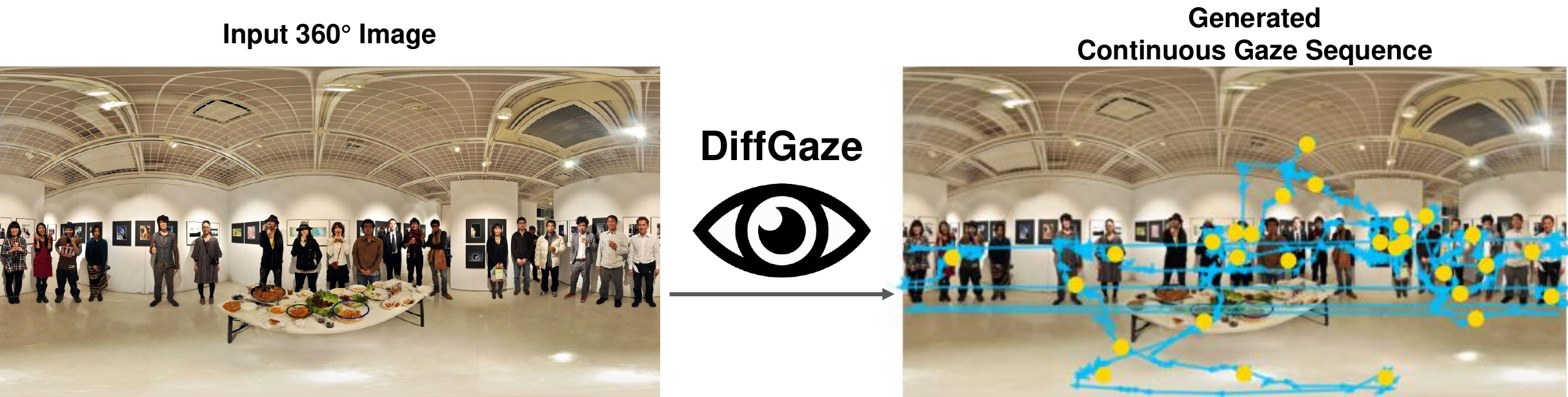}
\captionof{figure}{In stark contrast to scanpath prediction, which focuses on predicting sequences of discrete fixations (yellow), \textit{continuous gaze sequence generation} is the challenging task of predicting gaze data (blue) that resemble the data recorded using eye trackers. We present \textit{\methodName} -- the first method for continuous gaze sequence generation on 360$^\circ$ images. \methodName formulates the continuous gaze sequence generation task as a conditional diffusion process, with the image features as condition, and uses two transformers to capture the temporal and spatial dependencies of human gaze behaviour.}}

\makeatletter
\apptocmd{\@maketitle}{\insertfig}{}{}%
\makeatother

\begin{document}

\title{\methodName: A Diffusion Model for Continuous Gaze Sequence Generation on 360$^{\circ}$ Images}

\author{Chuhan~Jiao,~Yao~Wang,~Guanhua~Zhang,~Mihai~Bâce,~Zhiming~Hu,~Andreas~Bulling
\IEEEcompsocitemizethanks{\IEEEcompsocthanksitem 
Chuhan Jiao, Yao Wang, Guanhua Zhang, Zhiming Hu, and Andreas Bulling are with the Institute for Visualisation and Interactive Systems, University of Stuttgart, Germany. \protect E-mail: \{chuhan.jiao, yao.wang, guanhua.zhang, zhiming.hu, andreas.bulling\}@vis.uni-stuttgart.de.
Mihai~B\^{a}ce is with the Department of Computer Science, KU Leuven, Belgium. Part of this work was conducted while at the University of Stuttgart. \protect E-mail: mihai.bace@kuleuven.be.
}
\thanks{Manuscript submitted December 4, 2023.}}

\markboth{}%
{Jiao \MakeLowercase{\textit{et al.}}: \methodName: A Diffusion Model for Continuous Gaze Sequence Generation on 360$^{\circ}$ Images}

\maketitle

\begin{abstract}

We present \textit{\methodName}, a novel method for generating realistic and diverse continuous human gaze sequences on 360$^{\circ}$ images based on a conditional score-based denoising diffusion model.
Generating human gaze on 360$^{\circ}$ images is important for various human-computer interaction and computer graphics applications, e.g. for creating large-scale eye tracking datasets or for realistic animation of virtual humans. 
However, existing methods are limited to predicting discrete fixation sequences or aggregated saliency maps, thereby neglecting crucial parts of natural gaze behaviour.
Our method uses features extracted from 360$^{\circ}$ images as condition and uses two transformers to model the temporal and spatial dependencies of continuous human gaze.
We evaluate \methodName on two 360$^{\circ}$ image benchmarks for gaze sequence generation as well as scanpath prediction and saliency prediction.
Our evaluations show that \methodName outperforms state-of-the-art methods on all tasks on both benchmarks.
We also report a 21-participant user study showing that our method generates gaze sequences that are indistinguishable from real human sequences.
Taken together, our evaluations not only demonstrate the effectiveness of \methodName
but also point towards a new generation of methods that faithfully model
the rich spatial and temporal nature of natural human gaze behaviour.

\end{abstract}

\begin{IEEEkeywords}
Scanpath Prediction; Saliency Modelling; Eye Tracking; Gaze Behaviour Modelling; Eye Movement Synthesis 
\end{IEEEkeywords}

\section{Introduction}

With recent advances in camera technology, capturing high-resolution 360$^{\circ}$ images enables a new generation of immersive experiences in virtual reality (VR). 
This potential has led to rising consumer interest in adopting this new technology and growing research efforts in understanding how humans perceive and explore these 3D virtual environments~\cite{sitzmann2018saliency, sidenmark2019eye, rivu2021exploring}.
A particularly rich source of information on this exploratory process is visual attention, typically analysed in the form of gaze data collected using eye tracking.
While eye tracking has become more widely available and affordable \cite{kassner14_ubicomp,tonsen17_imwut} -- and is integrated into an ever-increasing number of VR headsets -- collecting gaze data, particularly at scale, remains tedious and time-consuming and is often not feasible at all.
This has triggered research into computational models of visual attention, i.e. models that predict human gaze on 360$^{\circ}$\ images without requiring special-purpose eye tracking equipment. 

Prior works on computational modelling of visual attention on 360$^{\circ}$\ images have focused on saliency~\cite{sitzmann2018saliency, chen2020salbinet360} or scanpath prediction~\cite{assens2017saltinet, martin2022scangan360, sui2023scandmm}.
Despite significant advances, both tasks still only tackle a simplified problem: While aggregated saliency maps do not require to model the temporal nature of human gaze behaviour, sequences of predicted discrete gaze fixations (scanpaths) are temporarily coarse and neglect rich gaze data between fixations.
As a result, neither of these tasks -- nor any existing method developed in the past to tackle them -- allow to faithfully model
the rich spatial and temporal nature of natural human gaze behaviour on 360$^{\circ}$\ images.

To address these limitations we propose \textit{\methodName} -- the first generative method to synthesise continuous human gaze sequences
on 360$^{\circ}$ images. 
\methodName is based on a score-based denoising diffusion model~\cite{tashiro2021csdi} that is conditioned on the features extracted from 360$^{\circ}$ images and uses two Transformers to model spatio-temporal human gaze behaviour.
We evaluate our method on continuous gaze sequence generation as well as scanpath prediction and saliency prediction on two datasets, Sitzmann~\cite{sitzmann2018saliency} and Salient360!~\cite{salient360rai2017dataset, salient360rai2017saliency}. 
The results show that our method achieves state-of-the-art performance on continuous gaze sequence generation and scanpath prediction, and outperforms other gaze sequence generation baselines in saliency prediction.
We also show that our model reproduces natural eye movement characteristics observable in humans, such as the mean number of saccades, mean saccade velocity, mean number of fixations, or mean fixation duration. 
Finally, through an online user study with 21 participants, we show that the gaze sequences generated by \methodName are practically indistinguishable from real human gaze behaviour.
Taken together, \methodName paves the way for a new family of generative methods that can faithfully model continuous human gaze behaviour.
As such, our method and the continuous gaze sequence generation task not only have the potential to unify long-standing, yet so far largely separated, efforts on saliency and scanpath prediction.
Our method also promises significant improvements in a range of applications that can directly build on it, such as animation of virtual humans.

\noindent
The specific contributions of our work are three-fold:
\begin{enumerate}[leftmargin=*, topsep=0pt, itemsep=0pt]
    \item We propose \methodName - the first generative method to synthesise continuous human gaze sequences on 360$^\circ$ images.
    \item  Through extensive evaluations, we show that our method achieves state-of-the-art performance on continuous gaze sequence generation and scanpath prediction.
    \item  Through an online user study with 21 participants, we show that the gaze sequences generated by \methodName are practically indistinguishable from real human gaze behaviour. 
\end{enumerate}

\section{Related Work}

Our work on continuous gaze sequence generation is related to previous work on 1) saliency modelling, 2) scanpath prediction, and 3) realistic eye movement generation.

\subsection{Saliency Modelling}

Saliency modelling aims at predicting the spatial distribution of human gaze fixations, also known as saliency map, on an image and has been extensively studied in the past few decades.
Traditional saliency prediction methods usually focus on predicting the saliency map of 2D images and can be classified into bottom-up and top-down approaches.
Bottom-up methods predict saliency map using the low-level image features such as intensity, colour, and orientation~\cite{itti1998model, cheng2015global}.
Top-down approaches use high-level features such as specific tasks and context information to predict saliency map~\cite{borji2012probabilistic, xu2016spatio}.
Recently, with the development of virtual reality, human visual saliency on 360$^\circ$ images has become an essential research topic in computer vision.
Specifically, Sitzmann et al. proposed to combine existing 2D saliency predictors with a central bias to generate saliency maps for 360$^\circ$ images~\cite{sitzmann2018saliency} while Chen et al. proposed a local-global bifurcated deep network for saliency prediction on 360$^\circ$ content~\cite{chen2020salbinet360}.
However, saliency prediction methods can only model the spatial distribution of human visual attention and ignore the temporal dynamics.
To solve this limitation, researchers started to focus on scanpath prediction that models the temporal sequence of gaze fixations.

\subsection{Scanpath Prediction}

Scanpath prediction aims to generate a discrete sequence of eye gaze fixations from a given image and has been extensively explored in vision research.
Early works mostly focus on generating scanpath from 2D images.
Specifically, Liu et al. extracted semantic features from the image and then applied a Hidden Markov Model to predict human scanpath~\cite{liu2013semantically}.
Bao et al. proposed a convolutional neural network to simultaneously predict the foveal saliency map and fixation duration by combining the inhibition of return with image content~\cite{bao2020human}.
In the rapidly evolving domain of virtual reality, the prediction of scanpaths on 360$^\circ$ images has emerged as a pivotal research topic~\cite{assens2017saltinet, martin2022scangan360, sui2023scandmm}. %
Martin et al. proposed to use a generative adversarial network combined with a dynamic time warping-based loss function to generate scanpaths for 360$^\circ$ content~\cite{martin2022scangan360} while Sui et al. presented a deep Markov model to generate scanpaths from 360$^\circ$ images by combining a semantics-guided transition function and a state initialisation strategy~\cite{sui2023scandmm}.
However, these prior methodologies for scanpath prediction primarily concentrate on generating discrete gaze fixation sequences, neglecting the continuous nature of human eye movements. In contrast, our work aims to predict continuous gaze sequences from 360$^\circ$ images rather than only generate discrete gaze fixations, thereby capturing a more realistic representation of human visual attention.%

\subsection{Eye Movement Generation}

Generating realistic eye movements has emerged as an important research topic in computer graphics and virtual reality given its relevance for several applications, such as virtual character animation or human attention analysis.
In early work, Lee et al. proposed to use an empirical model of saccades and a statistical model of eye-tracking data to generate eye movements~\cite{lee2002eyes} while Lan et al. presented a novel set of psychology-inspired generative models to synthesise eye movements in reading, verbal communication, and scene perception~\cite{lan2022eyesyn}.
Considering the strong correlation between eye and head movements, Sitzmann et al.~\cite{sitzmann2018saliency} and Rai et al.~\cite{salient360rai2017dataset} proposed to use head orientation as a proxy to eye gaze.
Hu et al. proposed to generate realistic eye movements in free-viewing situations from users' head movements and scene content~\cite{hu2019sgaze,hu2020dgaze} and to predict eye movements using task-related information in task-oriented situations~\cite{hu2021fixationnet,hu2021eye}.
Xu et al. proposed to predict eye movements in graphical user interface from users' mouse and keyboard inputs while Koulieris et al. used game state variables in video games to predict users' gaze positions~\cite{koulieris2016gaze}.
However, existing methods usually rely on other input modalities, such as head movements, mouse and keyboard inputs, to generate eye movements.
In stark contrast, we generate realistic and continuous gaze sequences only using the information available in 360$^\circ$ images.

\section{Method}
\label{sec:method}

\begin{figure*}
    \centering
    \includegraphics[width=\linewidth]{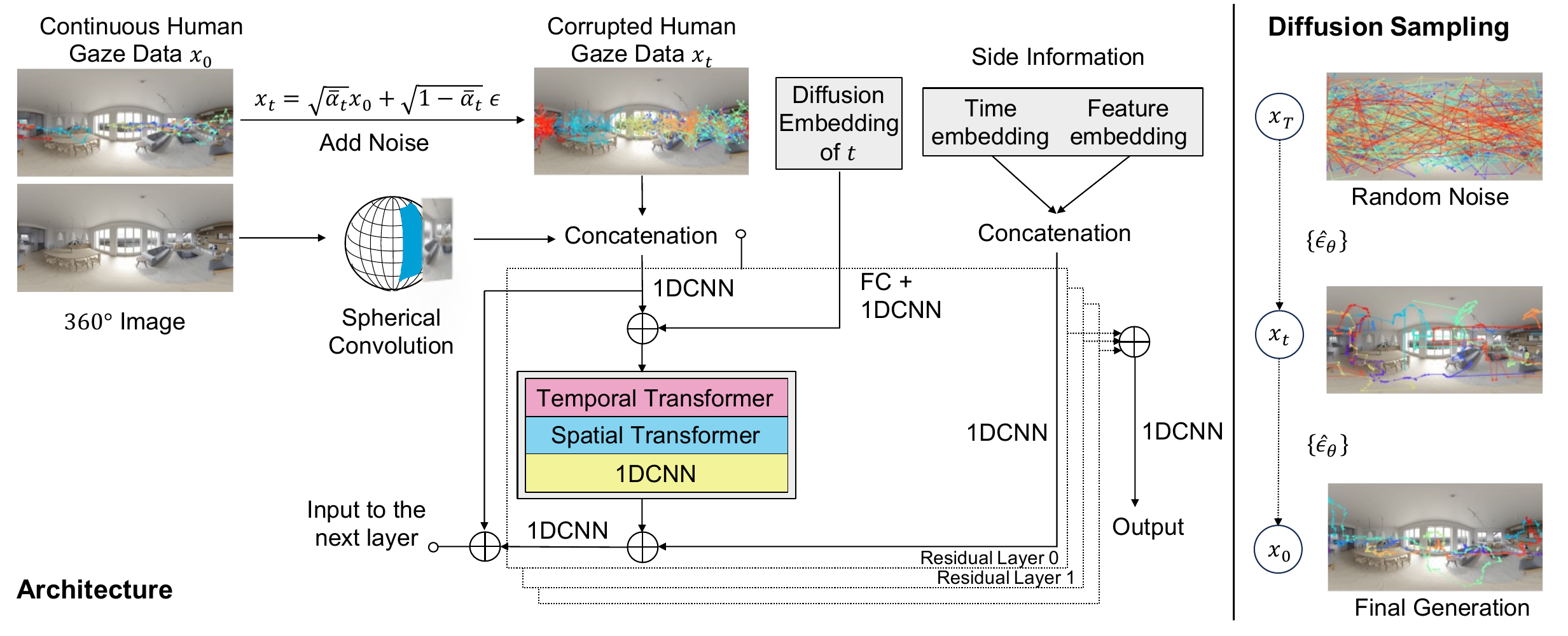}
    \caption{Overview of our proposed \methodName method.
    We cast continuous gaze sequence generation as a conditional diffusion task.
    Our model is trained to recover the original gaze trajectory from the corrupted, noisy data.
    The condition to guide this diffusion process includes a spherical convolution for the 360$^{\circ}$ image, and side information (time and feature embedding).
    We apply two Transformers to learn both the temporal and spatial attention.
    Please refer to the text for details about the architecture, diffusion process and the loss function.
    }
    \label{fig:method-overview}
\end{figure*}

An overview of our method is shown in ~\autoref{fig:method-overview}.
We formulate continuous gaze sequence generation on 360$^{\circ}$ images as a conditional diffusion task.
That is, \methodName is trained to denoise human gaze data corrupted by Gaussian noise, conditioned on the spherical convolution of a 360$^{\circ}$ image.
Our method uses two transformers to capture the temporal and spatial dependencies of continuous human gaze sequences.
During diffusion sampling, \methodName generates continuous gaze data from random noise and extracted image features.

\subsection{Problem Definition}
\label{sec:method-prob-def}
Eye tracking data are usually denoted as a sequence of two-dimensional locations $(i, j)$.
Projected on 360$^{\circ}$ images, these locations are represented by their latitude and longitude $(\phi,\lambda)$, where $\phi\in\left [ -\frac{\pi}{2},\frac{\pi}{2} \right ]$ and $\lambda\in\left [ -\pi,\pi \right ]$.
However, this representation introduces discontinuities at the image borders and leads to periodic values, causing problems during training.
For example, $\lambda$ and $\lambda+2\pi$ represent the same meridian, including the two image boarders ($\lambda =\pm 180^{\circ}$).
Thus, we instead project the image onto a unit sphere and represent each gaze point as a three-dimensional vector $(x, y, z)$ following~\cite{martin2022scangan360}:
\begin{equation}
    \label{eq:proj-sphere}
    x=cos(\phi)cos(\lambda), y=cos(\phi)sin(\lambda), z=sin(\phi)
\end{equation}
Hence, given a 360$^{\circ}$ image, we aim to generate a realistic series of gaze samples $X\in R^{3\times L}$.
$X$ consists of $L$ gaze points, where each point has three elements $(x,y,z)$.
The generated sequence is projected back to the latitude-longitude representation  $(\phi,\lambda)$ for evaluation and visualisation by
\begin{equation}
    \label{eq:proj-back}
    \phi=arctan2(z,\sqrt{x^2+y^2}), \lambda=arctan2(y,x)
\end{equation}

\subsection{Diffusion Model}
\label{sec:method-diffusion}
At its core, \methodName uses a denoising diffusion model that have recently shown impressive results in generative image tasks \cite{tashiro2021csdi}.
Diffusion models are probabilistic models, consisting of a forward and a reverse process.
The aim is to learn a distribution $p_{\theta}(X)$ to approximate the data distribution $q(X)$.
The \textbf{forward process} gradually adds Gaussian noise to the original data $X_0$ for $T$ timesteps.
At each step $t\in [1,T]$, the sampling of $X_t$ can be obtained via a Markov chain:
\begin{equation}
    q(X_t\mid X_{t-1})=\mathcal{N}(\sqrt{1-\beta _t}X_{t-1}; \beta _t\mathbf{I})
\end{equation}
where $\beta _t\in[0,1]$ is a scalar representing the noise level.
Using $\alpha_t=\prod_{i=1}^{t}(1-\beta_i)$, the noise distribution at any intermediate timestamp can be obtained by
\begin{equation}
    q(X_t\mid X_0)=\mathcal{N}(X_t; \sqrt{\alpha_t}X_0, (1-\alpha_t)\mathbf{I})
\end{equation}
Hence, $X_t$ is
\begin{equation}
    X_t=\sqrt{\alpha_t}X_0+(1-\alpha_t)\epsilon
\end{equation}
where $\epsilon\sim \mathcal{N}(0,\mathbf{I})$. 
In the \textbf{reverse process}, a generative model $\theta$ denoises $X_t$ to recover $X_0$ by the transition:
\begin{equation}
    \label{eq:reverse-ori}
    p_{\theta}(X_{t-1}\mid X_t)=\mathcal{N}(X_{t-1};\mu_\theta(X_t,t), \sigma_\theta(X_t,t)\mathbf{I})
\end{equation}
$\sigma_\theta(X_t,t)$ is fixed to a constant $\beta _t$ for easier optimisation.
$\mu_\theta(X_t,t)$ can be decomposed into the linear combination of $X_t$ and a denoising function $\epsilon_\theta(X_t,t)$, which is learnt via the following optimisation~\cite{ho2020denoising}:
\begin{equation}
    \underset{\theta}{min}\mathcal{L}(\theta)=\underset{\theta}{min}\mathbb{E}_{X_0\sim q(X_0), \epsilon \sim \mathcal{N}(0,\mathbf{I}),t}\left \| \epsilon - \epsilon_\theta(X_t,t) \right \|_{2}^{2}
\end{equation}

\subsection{Conditioning Mechanism for 360$^{\circ}$ Images}
\label{sec:method-condition}

Since objects in the image, their semantics and location, have a strong influence on human gaze behaviour, we use
the 360$^{\circ}$ image itself as a condition to the diffusion process.
More specifically, given that gaze points are projected to a spherical space (\autoref{eq:proj-sphere}), we propose a spherical embedding of 360$^{\circ}$ image as the condition in the reverse diffusion process.
We encode the semantic information of the image into $c \in R^{m\times 1}$ using the sphere convolutional neural network (S-CNN)~\cite{coors2018spherenet} that can handle space-varying distortions resulting from the equirectangular projection.
In S-CNN, we apply a CoordConv layer~\cite{liu2018intriguing} to extract the spatial embedding.
CoordConv adds extra coordinate channels to make convolutions more robust to coordinate transfromations, with few parameters and efficient computation.
After introducing the condition $c$ to the reverse process, the goal is to approximate the learnt distribution $p_{\theta}(X\mid c)$ to the genuine distribution $q(X\mid c)$.
Correspondingly, we extend the reverse process in \autoref{eq:reverse-ori} as
\begin{equation}
    p_{\theta}(X_{t-1}\mid X_t,c)=\mathcal{N}(X_{t-1};\mu_\theta(X_t,t\mid c), \sigma_\theta(X_t,t\mid c)\mathbf{I})
\end{equation}
with the training objective
\begin{equation}
    \underset{\theta}{min}\mathcal{L}(\theta)=\underset{\theta}{min} \mathbb{E}_{X_0\sim q(X_0), \epsilon \sim \mathcal{N}(0,\mathbf{I}),t}\left \| \epsilon - \epsilon_\theta(X_t,t\mid c)) \right \|_{2}^{2}
\end{equation}

\subsection{Implementation Details}
\label{sec:method-DiffGaze}

Our method builds on the DiffWave architecture~\cite{kong2020diffwave} that consists of multiple residual layers with bidirectional dilated convolutions.
We extend this architecture by introducing %
a two-dimensional attention mechanism in each residual layer to capture the temporal (along $L$) and spatial (among $x,y,z$) dependencies of the gaze trajectories.
We implement the attention mechanism using two one-layer Transformer encoders~\cite{tashiro2021csdi}:
One Transformer learns temporal dependencies for each feature, while the other captures spatial correlations at each timestamp.
At each diffusion step, we concatenate the noisy gaze sequence $X$ with the spherical image embedding as a condition $c$ and feed them into a convolutional layer.
In addition, we calculate two types of side information: 1) A 128-dimensional time embedding of the timestamps $S=\{S_{1:L}\}$ corresponding to the gaze trajectory~\cite{vaswani2017attention,zuo2020transformer}
\begin{equation}
    \begin{aligned}
    S_{embedding}(S_l)=(sin(S_l/\tau^{0/64}),...,
    sin(S_l/\tau^{63/64}),\\cos(S_l/\tau^{0/64}),..., 
    cos(S_l/\tau^{63/64}))
    \end{aligned}
\end{equation}
where $l\in [1,L]$, $\tau=10,000$, 
as well as 2) a 16-dimensional categorical embedding of the three features $(x,y,z)$~\cite{tashiro2021csdi}.
We concatenate both embeddings, use them as input to a 1DCNN layer, and add them to the other embeddings (see \autoref{fig:method-overview}).

\section{Experiments}
\label{sec:exp}

\subsection{Datasets}
Most existing scanpath and saliency datasets for 360$^{\circ}$ do not offer real gaze data captured using an eye tracker to train and evaluate attention models.
Two noteworthy exceptions are 
Sitzmann \cite{sitzmann2018saliency} and Salient360! \cite{salient360rai2017dataset, salient360rai2017saliency}.
The Sitzmann dataset contains 22 images with 1,980 gaze trajectories from 169 participants, recorded at a sampling rate of 120\,Hz for 30 seconds per image. 
Salient360! contains binocular gaze data of between 40\,-\,42 observers looking at 85 images.
Gaze data was recorded at a sampling rate of 60\,Hz for 25 seconds per image. 
Following previous works on scanpath prediction~\cite{martin2022scangan360,sui2023scandmm}, we used the same 19 images from the Sitzmann dataset for training and the remaining three for testing.
We used the entire Salient360! dataset as a test set for cross-dataset evaluation.

In eye tracking, the number of gaze samples should (roughly) be the sampling frequency multiplied by the recording duration. 
We discarded recordings with fewer gaze samples than the minimum number of gaze samples. For Sitzmann, we set the minimum number to 3,481, corresponding to 29 seconds of viewing time. For Salient360!, we set the minimum number to 1,441, corresponding to 24 seconds of viewing time. 

We opted to generate continuous gaze sequences at a sampling frequency of 30\,Hz
as commonly offered by commercial eye trackers.
To this end, we first downsampled the data from both datasets to 30\,Hz.
To fix the output size for the diffusion model, we then truncated the downsampled data to 871 samples for Sitzmann and 721 samples for Salient360!.

\subsection{Experimental Setup} \label{sec:exp_setup}
We used the downsampled Sitzmann dataset to train our model.
Similar to \cite{sui2023scandmm}, we resized all images to a resolution of (128, 256) before training.
We adapted the diffusion process parameters from CSDI \cite{tashiro2021csdi}
and set the noise level range from {$0.0001$} to {$0.5$}.
We also used a quadratic schedule to update the noise level at each diffusion step, following \cite{tashiro2021csdi, song2020denoising, nichol2021improved} to enhance the quality of the generated samples.
We set the total diffusion step {$T = 200$}, the batch size as 16, and the learning rate to 0.001. We used the Adam optimiser to train our model for 500 epochs, and reduce the learning rate by a factor of 0.1 at epoch 375 and epoch 450.

\textbf{Baselines.} To the best of our knowledge, \methodName is the first method for generating 30\,Hz continuous gaze sequences on 360$^{\circ}$ images. The closest methods to ours are ScanGAN360 \cite{martin2022scangan360} and ScanDMM \cite{sui2023scandmm} but they were designed for generating 1\,Hz gaze sequences to simulate the human scanpaths. We adapted these methods by changing their output resolutions and by retraining them using our settings. We used the same hyperparameters as given in the original papers in retraining.

\begin{table*}[t]
    \caption{Quantitative evaluation of continuous gaze sequence generation methods on Sitzmann and Salient360! dataset in terms of Levenshtein distance~(LEV), Dynamic Time Warping~(DTW), mean absolute error~(MAE), and root mean squared error~(RMSE). Best results are shown in \textbf{bold}.}
    \centering
    \begin{tabular}{llllllllll}
    \toprule
    \multirow{2}{*}{\textbf{Dataset}} & \multirow{2}{*}{\textbf{Method}} & \multicolumn{2}{c}{\textbf{LEV~$\downarrow$}} & \multicolumn{2}{c}{\textbf{DTW~$\downarrow$}} & \multicolumn{2}{c}{\textbf{MAE~$\downarrow$}} & \multicolumn{2}{c}{\textbf{RMSE~$\downarrow$}} \\
    & & \textit{mean} & \textit{best} & \textit{mean} & \textit{best} & \textit{mean} & \textit{best} & \textit{mean} & \textit{best} \\
    \midrule
    \multirow{4}{*}{Sitzmann \cite{sitzmann2018saliency}} &
    Human & 1,167$^\dag$ & 992$^\dag$ 
    & 1,708,925$^\dag$ & 944,914$^\dag$
    & 1,587$^\dag$ & 1,123$^\dag$ 
    & 2,418$^\dag$ & 1,813$^\dag$ \\
    \cmidrule{2-10}
    & ScanGAN360~\cite{martin2022scangan360} & 1,406 & 1,327 
    & 2,084,017 & 1,709,852	
    & 1,795 & 1,422 
    & 2,532 & 2,051 \\
    & ScanDMM~\cite{sui2023scandmm} & 1,293 & 1,161 
    & 2,139,311 & 1,467,909
    & 1,710 & 1,279 
    & 2,519 & 1,966 \\
    & \methodName~(Ours) & \textbf{1,272} & \textbf{1,148} 
    & \textbf{1,785,109} & \textbf{1,163,822}
    & \textbf{1,623} & \textbf{1,195}
    & \textbf{2,365} & \textbf{1,810} \\\midrule

    \multirow{4}{*}{Salient360! \cite{salient360rai2017dataset, salient360rai2017saliency}} &
    Human & 1,060$^\dag$ & 928$^\dag$
    & 360,523$^\dag$ & 215,671$^\dag$
    & 402$^\dag$ & 285$^\dag$ 
    & 587$^\dag$ & 434$^\dag$ \\
    \cmidrule{2-10}
    & ScanGAN360~\cite{martin2022scangan360} & 1,163 & 1,084
    & 421,229 & 337,640
    & 435 & 344
    & 602 & 478 \\
    & ScanDMM~\cite{sui2023scandmm} & 1,092 & \textbf{951} 
    & 441,725 & 285,542
    & 421 & 301
    & 611 & 456 \\
    & \methodName~(Ours) & \textbf{1,079} & 957
    & \textbf{380,857} & \textbf{245,206}
    & \textbf{413} & \textbf{298}	
    & \textbf{591} & \textbf{439} \\
    \bottomrule
    \end{tabular}\\
    \vspace{2pt}
    \footnotesize{$^\dag$ Gaze sequences are not compared with themselves}
    \label{table:raw_quan} 
\end{table*}

\begin{figure*}[t]
    \centering
    \includegraphics[width=\linewidth]{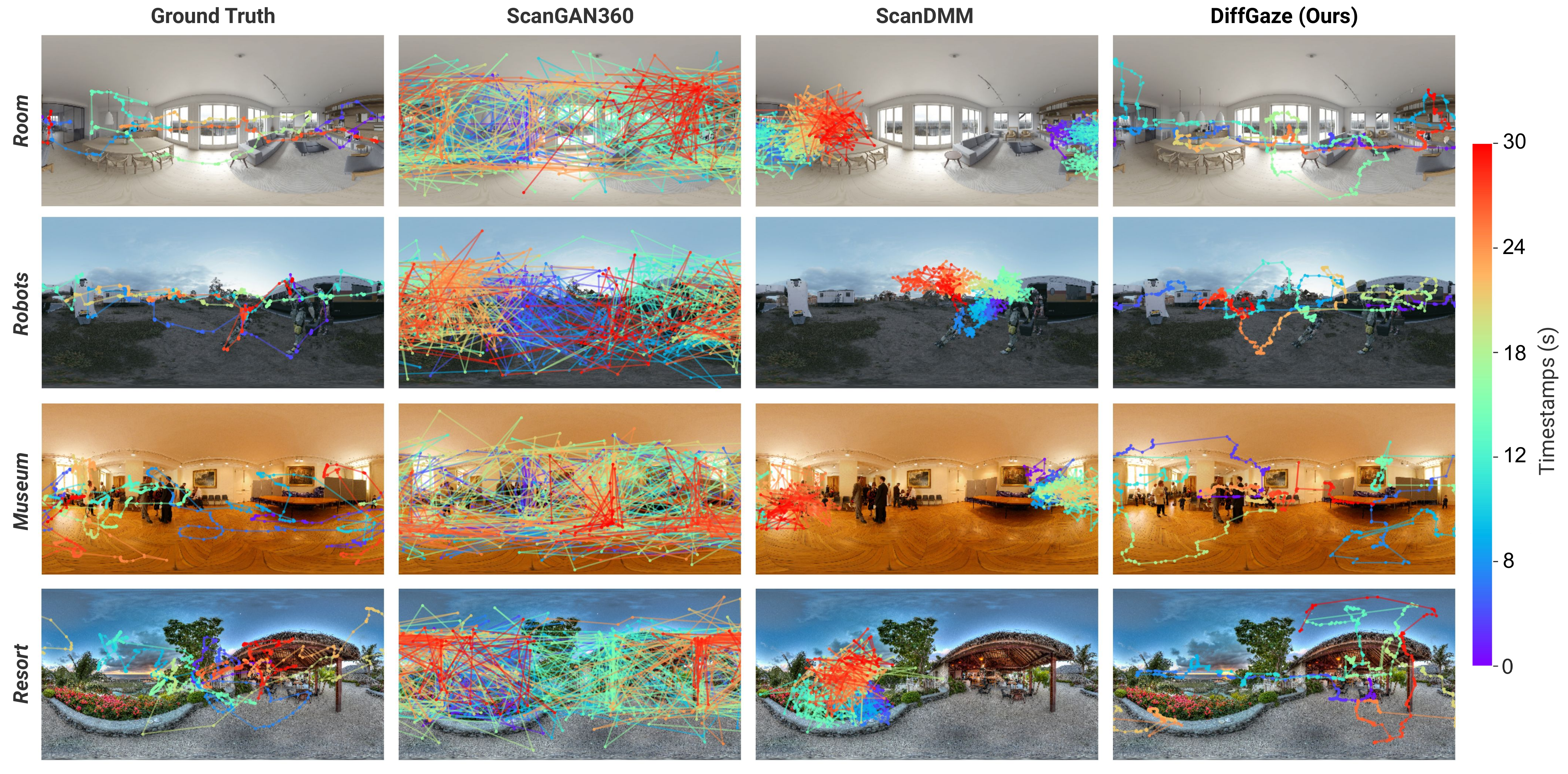}
    \caption{Qualitative comparison of continuous gaze data generation models in four scenes. From left to right: gaze samples from a human observer, generated 30\,Hz eye movement sequences from the ScanGAN360 method, ScanDMM, and our proposed model. From top to bottom: the Room and the Robots from the Sitzmann dataset, the Museum and Resort from the Salient360! dataset.}
    \label{fig:experiment_rawgaze}
\end{figure*}

\subsection{Continuous Gaze Sequence Generation} \label{sec:gaze_seq_exp}
We conducted an experiment to assess the quality of the gaze sequences generated by our method and the baselines at 30 Hz, compared to the human ground truth. All methods are generative models that produce diverse gaze sequences, therefore we generated 100 sequences per method for each image to reduce the sampling bias.

\textbf{Evaluation Metrics. } We evaluated the generated gaze sequences using four time-series metrics commonly used in scanpath prediction and gaze data super-resolution: Levenshtein distance (LEV), dynamic time warping (DTW), mean absolute error (MAE), and root mean square error (RMSE) \cite{sui2023scandmm, martin2022scangan360, jiao23_supreyes}. These metrics assess the temporal alignment, spatial distance, and overall deviation between two gaze sequences, with lower values indicating better performance. All gaze locations were converted to image space for metric calculation. For the Sitzmann dataset, metrics were calculated on the original image size due to uniform resolution. For the Salient360! dataset, metrics were calculated on the provided saliency maps’ size (1024, 2048) due to varying image sizes.

Given that each image has multiple human ground truth gaze sequences and multiple generated gaze sequences, we report two numbers for each metric: \textit{best} and \textit{mean}, following prior work \cite{wang23_tvcg}. 
These numbers capture two important aspects of the quality of the generated sequences. The best number reflects how realistic the generated gaze sequence is, i.e., if there exists a human gaze sequence that is close to the generated one. We computed the best number as follows: For each generated sequence, we computed each metric to all ground truth sequences and recorded the best score for each metric. We then averaged the best scores over all generated sequences for each image and reported the result.
In contrast to \textit{best}, the \textit{mean} number reflects how representative the generated gaze sequence is, i.e., how well it captures the average behaviour of all human gaze sequences. We computed the mean number as follows: For each generated sequence, we computed each metric with respect to all ground truth sequences and recorded the average score for each metric. Then we averaged the means over all generated sequences and reported the result.

To compare our results with human performance, we also computed a human baseline \cite{xia2019predicting_human} for each metric. For each image, we selected one ground truth sequence and computed the metrics with the remaining ground truth sequences. We recorded the best score for each metric and averaged them over all ground truth sequences for each image. We then calculated the mean over all images for the human baseline. 

\textbf{Quantitative Results.} Table \ref{table:raw_quan} shows the results on both datasets. \methodName demonstrates superior performance on the Sitzmann dataset across all metrics and outperforms other baselines on three out of four metrics on Salient360!. This shows our method's effectiveness in generating continuous gaze sequences that closely resemble human behaviour.

\textbf{Qualitative Results.} \autoref{fig:experiment_rawgaze} presents examples of the generated gaze sequences. The gaze samples produced by ScanGAN360 and ScanDMM did not resemble the human eye behaviour of exploring 360$^{\circ}$ images. In particular, ScanGAN360 created many rapid eye movements across the images, unlike the human ground truth. The samples produced by ScanDMM concentrate on a few large regions on the images. This suggests that these two methods do not capture the relation between fixation and saccade in human eye movements. On the other hand, \methodName shows similar behaviours to the human ground truth.  Additional qualitative results can be found in the supplementary material.

We also observed that visual inspection of human gaze sequences reveals significant variation in gaze patterns, even when viewing the same image. %
This variability results in poor performance of human agreement on existing scanpath metrics and time-series metrics (see \autoref{table:raw_quan}). 
Therefore, these metrics’ comparison with human agreement may not fully reflect the model’s performance. 

\textbf{User Study.} To gain a more nuanced understanding of model performance, we designed a user study, following prior work in scanpath prediction \cite{wang23_tvcg}, that involved users in rating visualisations of different continuous gaze trajectories. The user study was structured into three stages:
\begin{itemize}[leftmargin=*, topsep=0pt, itemsep=0pt]
    \item \textit{Training}: we randomly picked three 360$^\circ$ images from both dataset. For each image, we showed 10 randomly chosen human gaze sequences to familiarise participants with the visual appearance of real human gaze data.
    \item \textit{Rating}: Participants were then asked to rate the realism of gaze sequences on a scale of 1 to 10, where 1 meant highly unrealistic, and 10 highly realistic. We randomly selected 20 images from both datasets. For each image, we randomly showed one real human gaze sequence, one sequence from our method, ScanDMM, and ScanGAN360, respectively. To minimise any potential positional bias, the presentation order of the four visualisations was randomised.
    \item \textit{Validation}: We randomly selected 10 images from both datasets. For each image, we randomly picked four real human gaze sequences and asked participants to rate them in the same way as before. We then computed the average rating for each participant. We discarded all participants with an average rating lower than five.
\end{itemize}

\begin{figure}[t]
    \centering
    \includegraphics[width=\linewidth]{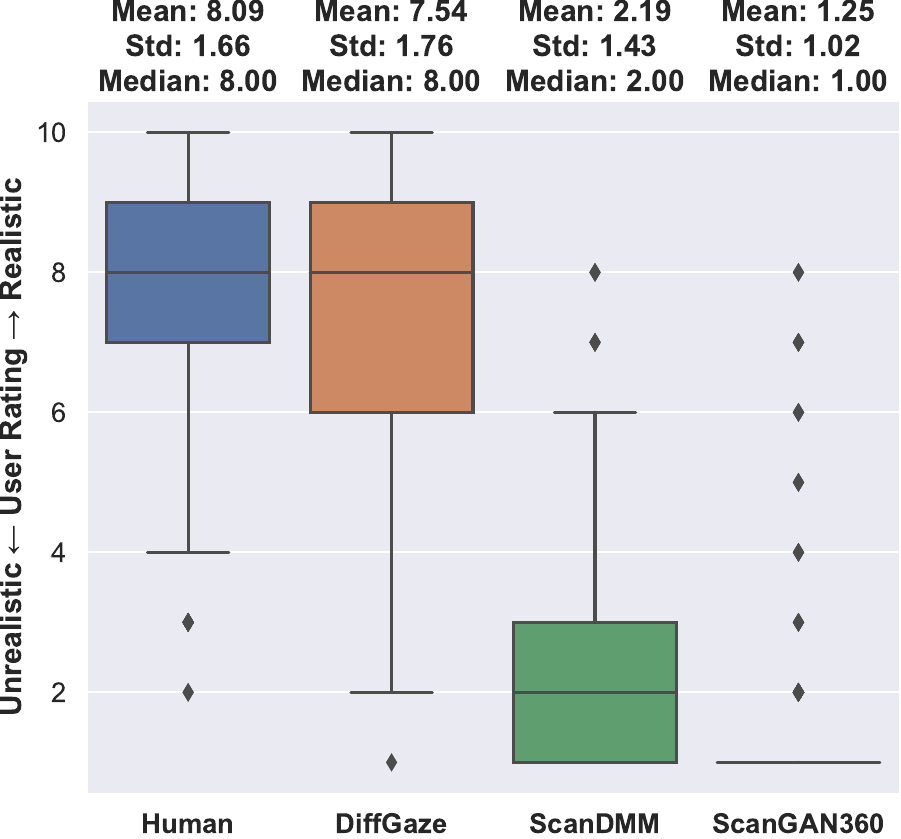}
    \caption{User ratings of the realism of gaze sequences generated by \methodName, ScanDMM, and ScanGAN360 (1: highly unrealistic, 10: highly realistic).}
    \label{fig:user_study}
\end{figure}

We recruited 21 participants (nine females and 12 males) from different universities for the study, 15 of them were familiar with gaze data or had participated in eye tracking studies.
The study was done online. No personal information of participants was collected. 
\autoref{fig:user_study} shows the average user ratings for realism achieved by the three different methods. 
The average rating of \methodName (7.54) is significantly higher than those of ScanDMM (2.19, t=63, p\textless0.001) and ScanGAN360 (1.25, t=93, p\textless0.001).  
Although the average of \methodName is slightly lower than the humans (7.54 vs 8.09), the median (8) is the same. This indicates that in many cases the generated continuous gaze sequences of our method are indistinguishable from real human sequences. 

\textbf{Analysis of Eye Movements.} 
We further evaluated \methodName by comparing the fixation and saccade statistics of the generated gaze sequences with human ground truth. We used a velocity-based saccade detection algorithm \cite{engbert2016evaluation} to segment the gaze sequences in spherical coordinates into fixations and saccades, with a velocity threshold $\lambda = 2$. %

\autoref{table:raw_stat} shows the results of this analysis. In terms of saccade statistics, \methodName and ScanGAN360 yield similar mean saccade velocities as human observers in both datasets. However, ScanDMM and ScanGAN360 generate more saccades than human observers, generating more frequent attention shifts. \methodName achieves a comparable number of saccades as human observers, suggesting more natural gaze behaviour. 
We can also see that ScanDMM and ScanGAN360 generate more fixations than human observers, with shorter mean fixation durations. 
\methodName produces more realistic fixation statistics in the Salient360! dataset, with a similar number and duration of fixations as human observers. On the Sitzmann dataset, \methodName generates slightly fewer fixations with slightly longer duration than human observers, but is still closer to the ground truth than the baselines. Taken together, these results show that \methodName can generate gaze sequences with characteristics that are highly similar to real human gaze behaviour.

\begin{table*}[t]
    \caption{Eye movement statistics on Sitzmann and Salient360! dataset. Numbers closest to human statistic are marked in bold} %
    \centering
    \begin{tabular}{llllll}
\toprule
    \textbf{Dataset} & \textbf{Method} & \textbf{Mean saccade} & \textbf{Mean saccade} & \textbf{Mean fixation} & \textbf{Mean fixation} \\ 
    & & \textbf{number} & \textbf{velocity ($^\circ$/s)} & \textbf{number} & \textbf{duration (s)} \\ 
\midrule
\multirow{4}{*}{Sitzmann \cite{sitzmann2018saliency}}
    & Human & 47.73 ± 9.86  & 224.23 ±  24.66 & 41.11 ± 7.29 & 0.69 ± 0.13 \\ 
    \cmidrule{2-6}
    & ScanDMM~\cite{sui2023scandmm} & 103.28 ± 32.85  & 192.12 ± 42.52 & 60.89 ± 16.45 & 0.46 ± 0.16  \\
    & ScanGAN360~\cite{martin2022scangan360} & 154.93 ± 14.46 & 217.24 ± 6.06 & 91.22 ± 9.82 & 0.26 ± 0.03  \\
    & \methodName (Ours) & \textbf{41.50 ± 9.11} & \textbf{220.97 ± 28.05}	 & \textbf{32.80 ± 6.91} & 0.88 ± 0.22  \\
   
    \midrule
    
    \multirow{4}{*}{Salient360! \cite{salient360rai2017dataset, salient360rai2017saliency}} & Human & 44.05 ± 11.70 & 209.01 ±  24.73 & 30.94 ± 7.25 & 0.79 ± 0.30  \\ 
    
    \cmidrule{2-6}
    & ScanDMM~\cite{sui2023scandmm} & 91.34 ± 29.46 & 193.54 ± 46.33 & 54.68 ± 15.45 & 0.44 ± 0.15 \\
    & ScanGAN360~\cite{martin2022scangan360} &  144.32 ± 23.41 & 216.13 ± 7.62 & 84.80 ± 13.25 & 0.29 ± 0.06 \\
    & \methodName (Ours) & \textbf{35.94 ± 8.35} & \textbf{219.76 ± 29.85} & \textbf{28.58 ± 6.42} & \textbf{0.84 ± 0.21}  \\
\bottomrule
    \end{tabular}
    \label{table:raw_stat}
\end{table*}

\subsection{Scanpath Prediction} \label{sec:scanpath_prediction}

\begin{table*}[t]
    \caption{Quantitative evaluation of scanpath prediction methods on Sitzmann and Salient360! dataset in terms of Levenshtein distance~(LEV), Dynamic Time Warping~(DTW), and Recurrence~(REC). Best results are shown in \textbf{bold}.}
    \centering
    \begin{tabular}{llllllll}
    \toprule
    \multirow{2}{*}{\textbf{Dataset}} & \multirow{2}{*}{\textbf{Method}} & \multicolumn{2}{c}{\textbf{LEV~$\downarrow$}} & \multicolumn{2}{c}{\textbf{DTW~$\downarrow$}} & \multicolumn{2}{c}{\textbf{REC~$\uparrow$}} \\
    & & \textit{mean} & \textit{best} & \textit{mean} & \textit{best} & \textit{mean} & \textit{best} \\
    \midrule
    \multirow{11}{*}{Sitzmann \cite{sitzmann2018saliency}} &
    Human & 55.82$^\dag$ & 44.89$^\dag$
    & 73,751$^\dag$ & 43,655$^\dag$
    & 0.108$^\dag$ & 0.541$^\dag$\\
    \cmidrule{2-8}
    &Human~(1\,Hz) & 43.64 & 35.51
    & 35,712 & 24,088 
    & 0.459 & 1.393\\
    & ScanGAN360~(1\,Hz)~\cite{martin2022scangan360} & \textbf{54.29} & 41.92
    & 70,764 & 46,247 
    & 0.044 & 0.361 \\
    & ScanDMM~(1\,Hz)~\cite{sui2023scandmm} & \textbf{54.29} & 41.25
    & 71,250 & 45,452
    & 0.049 & 0.375 \\
    & \methodName~(1\,Hz, Ours) & 54.88 & 41.45 
    & 71,861 & 45,466
    & 0.046 & 0.378 \\
    \cmidrule{2-8}
    & Saltinet~\cite{assens2017saltinet} & 64.73 & 54.35
    & 89,722 & 66,610 
    & 0.022 & 0.249 \\
    & DeepGaze III~\cite{kummerer2022deepgaze} & 59.48 & 51.94 
    & 78,083 & 57,813
    & 0.071 & 0.407 \\
    & CLE~\cite{boccignone2020look} & 63.01 & 53.63	
    & 83,846 & 46,037
    & \textbf{0.091} & \textbf{1.313} \\
    & ScanGAN360~\cite{martin2022scangan360} & 60.35 & 50.73
    & 74,644 & 50,635 
    & 0.037 & 0.302 \\
    & ScanDMM~\cite{sui2023scandmm} & 57.22 & 42.80
    & 88,513 & 52,559
    & 0.049 & 0.475 \\
    & \methodName~(Ours) & 55.84 & \textbf{39.40} 
    & \textbf{69,381} & \textbf{44,065}
    & 0.043 & 0.439 \\

    \midrule

    \multirow{7}{*}{Salient360! \cite{salient360rai2017dataset, salient360rai2017saliency}} &
    Human & 99.40 $^\dag$ & 76.44$^\dag$
    & 28,657$^\dag$ & 15,998$^\dag$
    & 0.092$^\dag$ & 0.479$^\dag$ \\
    \cmidrule{2-8}
    & Saltinet~\cite{assens2017saltinet} & 105.88 & 55.45 
    & 44,626 & 25,623
    & 0.009 & 0.177 \\
    & DeepGaze III~\cite{kummerer2022deepgaze} & 100.87 & 83.69
    & 33,390 & 25,573
    & 0.041 & 0.255 \\
    & CLE~\cite{boccignone2020look} & 100.11 & 79.06 
    & 32,535 & 16,020
    & \textbf{0.088} & \textbf{1.046} \\
    & ScanGAN360~\cite{martin2022scangan360} & \textbf{94.62} & 53.79
    & 25,281 & 15,261
    & 0.036 & 0.324 \\
    & ScanDMM~\cite{sui2023scandmm} & 96.71 & 45.26
    & 32,674 & 14,550
    & 0.045 & 0.552 \\
    & \methodName\,(Ours) & 98.85 & \textbf{42.20}
    & \textbf{25,238} & \textbf{12,641}
    & 0.038 & 0.535 \\
    \bottomrule
    \end{tabular}\\
    \vspace{2pt}
    \footnotesize{$^\dag$ Scanpaths are not compared with themselves}
    \label{table:scanpath_quan} 
\end{table*}
\begin{figure*}[t]
    \centering
    \includegraphics[width=\linewidth]{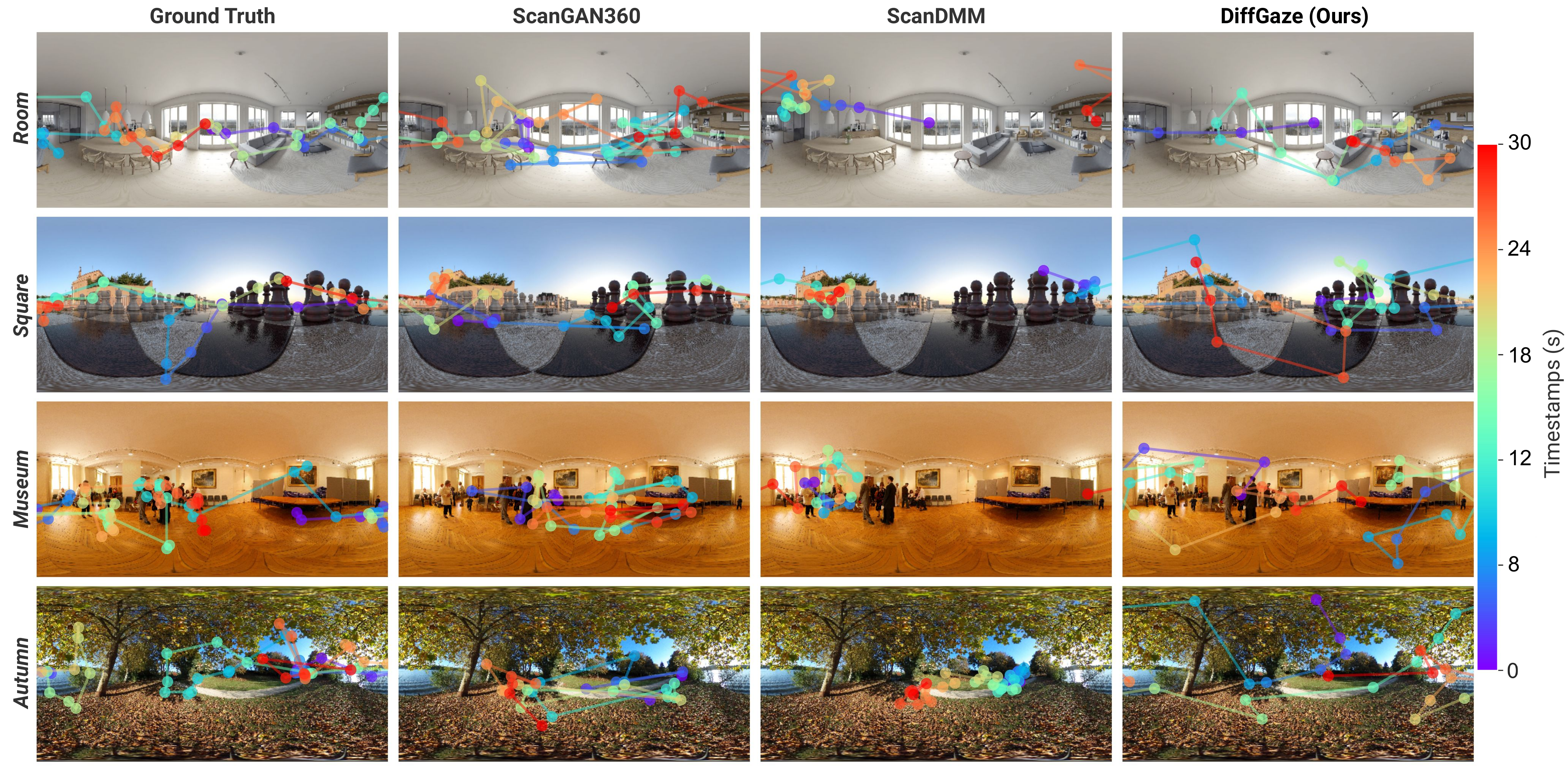}
    \caption{Qualitative comparison to scanpath prediction models in four scenes. From left to right: scanpaths obtained by a human observer, generated 30\,Hz scanpaths obtained by ScanGAN360, ScanDMM, and the proposed model. From top to bottom: the Room and the Square from Sitzmann dataset, the Museum and Autumn from Salient360! dataset.}
    \label{fig:experiment_scanpath}
\end{figure*}

Given that fixations and their durations are readily encoded in gaze sequences, we also evaluated \methodName on scanpath prediction. 
We used the scanpaths obtained from our gaze sequence extraction procedure (Section \ref{sec:gaze_seq_exp}) as ground truth on the Sitzmann dataset, given it does not provide them originally. We discarded fixations shorter than 150 ms, following \cite{sitzmann2018saliency} and MIT1003 \cite{Judd_2009}. The remaining fixations formed the scanpaths for our analyses. For the Salient360! dataset, we used the provided scanpaths as ground truth.

\textbf{Baselines.} In addition to the two baselines trained on continuous gaze sequences, we also compared our method with the state-of-the-art scanpath prediction methods, Saltinet \cite{assens2017saltinet}, DeepGaze III \cite{kummerer2022deepgaze}, CLE \cite{boccignone2020look}, and ScanGAN360 \cite{martin2022scangan360} and ScanDMM \cite{sui2023scandmm} which trained on 1\,Hz gaze sequences. For Saltinet, DeepGaze III, ScanGAN360 (1\,Hz), and ScanDMM (1\,Hz), we used the pre-trained weights provided by the authors. Since CLE generates scanpath on given saliency maps, we applied BMS360 \cite{lebreton2018gbvs360} to obtain the saliency maps. For CLE, Saltinet, and DeepGaze III, the number of generated fixations has to be pre-defined. Therefore, we set the number to the mean number of fixations of each dataset for these methods.

In addition, to link with 1\,Hz gaze sequence prediction, we trained a 1\,Hz \methodName model using the processed 1\,Hz Sitzmann dataset provided by ScanDMM authors \cite{sui2023scandmm} with the same setting as described in \ref{sec:exp_setup}. Moreover, we compared the 1\,Hz human ground truth of the Sitzmann dataset with the ground truth scanpaths extracted from human gaze sequences.

\textbf{Evaluation Metrics.} Following \cite{sui2023scandmm}, we used Levenshtein distance (LEV), dynamic time warping (DTW), and recurrence measure (REC) as evaluation metrics. REC measures the proportion of gaze points that are close to each other in two scanpaths. We opted for two degrees of visual angle as the threshold following \cite{salient360rai2017saliency}. All metrics and the human baseline \cite{xia2019predicting_human} were computed as described in Section \ref{sec:gaze_seq_exp}. 

\textbf{Quantitative Results.}  \autoref{table:scanpath_quan} shows 1Hz human gaze sequences exhibit superior metrics compared to scanpaths extracted from continuous human gaze sequences.
This suggests that the scanpath prediction task is oversimplified by downsampling the human gaze sequence to 1\,Hz. 
Consequently, on the Sitzmann dataset, the models trained with data have higher upper-bound performance than the actual human scanpath. Therefore, 1\,Hz models have overall good performance. For a fair comparison, we excluded the 1\,Hz models from other evaluations. However, it is worth noting that, even if generating continuous human gaze data is a more difficult task compared with 1\,Hz data generation, the scanpaths extracted from \methodName still outperform those 1\,Hz models on DTW, best REC, and best LEV. And it rivals those 1\,Hz models in other metrics. 

\methodName also shows promising results compared to scanpath prediction models and other baselines on both datasets. On the Sitzmann dataset, which has only three test images, \methodName achieves the best results and reaches human performance on LEV and DTW. On Salient360!, which has 85 test images, \methodName outperforms all the baselines and matches human performance on LEV and DTW, showing its strong generalisation ability in cross-dataset evaluation. Interestingly, \methodName has slightly higher LEV and DTW scores than human agreement, which suggests that \methodName produces more consistent scanpaths than human observers. As for REC, CLE has the best performance on both datasets, because REC only measures the spatial similarity of fixations and ignores the temporal aspect. CLE generates scanpaths from saliency maps, which gives it an advantage in capturing salient regions. However, \methodName still shows comparable performance to models without saliency prior, demonstrating its ability to generate spatially accurate fixations.

\textbf{Qualitative Results.} Figure \ref{fig:experiment_scanpath} shows example scanpaths obtained from generated continuous gaze sequences of \methodName, ScanGAN360, and ScanDMM. It can be observed that ScanDMM tends to produce fixations that are concentrated within a limited area, thereby failing to replicate the exploratory nature of human gaze. Conversely, the scanpaths generated by our method exhibit greater resemblance to the human ground truth in terms of the number, location, and sequence of fixations. For example, in the \textit{Museum} scene, our method successfully emulates the shift in human attention between the crowd on the left and the painting on the right. While the scanpaths produced by ScanGAN360 display a pattern akin to ours, \autoref{table:raw_stat} reveals that the average duration of fixations for ScanGAN360 is significantly reduced compared to human fixations. In contrast, the distribution of fixation durations for \methodName aligns closely with that of humans. Overall, the scanpaths derived from the gaze sequences generated by \methodName demonstrate a higher degree of similarity to human scanpaths compared to the other two baseline methods.

\subsection{Saliency Prediction}
The gaze sequence and scanpath evaluation focused on the temporal perspective. 
To better evaluate the generated gaze sequence spatially, we evaluated the different gaze sequence generation approaches, \methodName, ScanGAN360 \cite{martin2022scangan360}, and ScanDMM \cite{sui2023scandmm} on the saliency prediction task. We used each method to generate 1,000 30\,Hz gaze sequences and applied the eye event detection algorithm mentioned in Section \ref{sec:gaze_seq_exp} to obtain gaze fixations. Similar to \autoref{sec:scanpath_prediction}, we filtered out all the fixations with duration less than 150\,ms. 
We then used the script provided by Sitzmann et al. \cite{sitzmann2018saliency} to convert the fixation maps into continuous saliency maps by applying a Gaussian filter with a standard deviation of 1\textdegree\ of visual angle.

\textbf{Baselines.} Our method was trained to generate continuous gaze data and was not optimised for saliency prediction while existing deep learning saliency prediction methods on 360$^\circ$ images are trained on these two datasets. For a fair comparison, we did not compare to deep learning saliency prediction baselines. Instead, we report results using BMS360 \cite{lebreton2018gbvs360} and GBVS360 \cite{lebreton2018gbvs360}, which are two bottom-up saliency models that perform well
compared to the state-of-the-art methods in previous saliency prediction works \cite{sui2023scandmm, monroy2018salnet360, chao2018salgan360}. 

\textbf{Evaluation Metrics.}
We report performance using five common metrics: Area under the ROC Curve (AUC), Normalised Scanpath Saliency (NSS), Similarity (SIM), Pearson’s Correlation Coefficient (CC), and Kullback-Leibler divergence (KL). All metrics were calculated on image size (4096, 8192) for Sitzmann and (1024, 2048) for Salient360!, respectively.

\textbf{Quantitative Results.}
\autoref{table:sitzmann_saliency} shows the results for saliency prediction. Our method achieves superior performance over ScanDMM and ScanGAN360 on all evaluation metrics on the Sitzmann dataset. On Salient360!, our method outperforms the two baselines on four out of five metrics, except for KL. Moreover, \methodName ranks second in AUC on both datasets and improves the AUC of GBVS360.

\begin{figure*}[t]
    \centering
    \includegraphics[width=\linewidth]{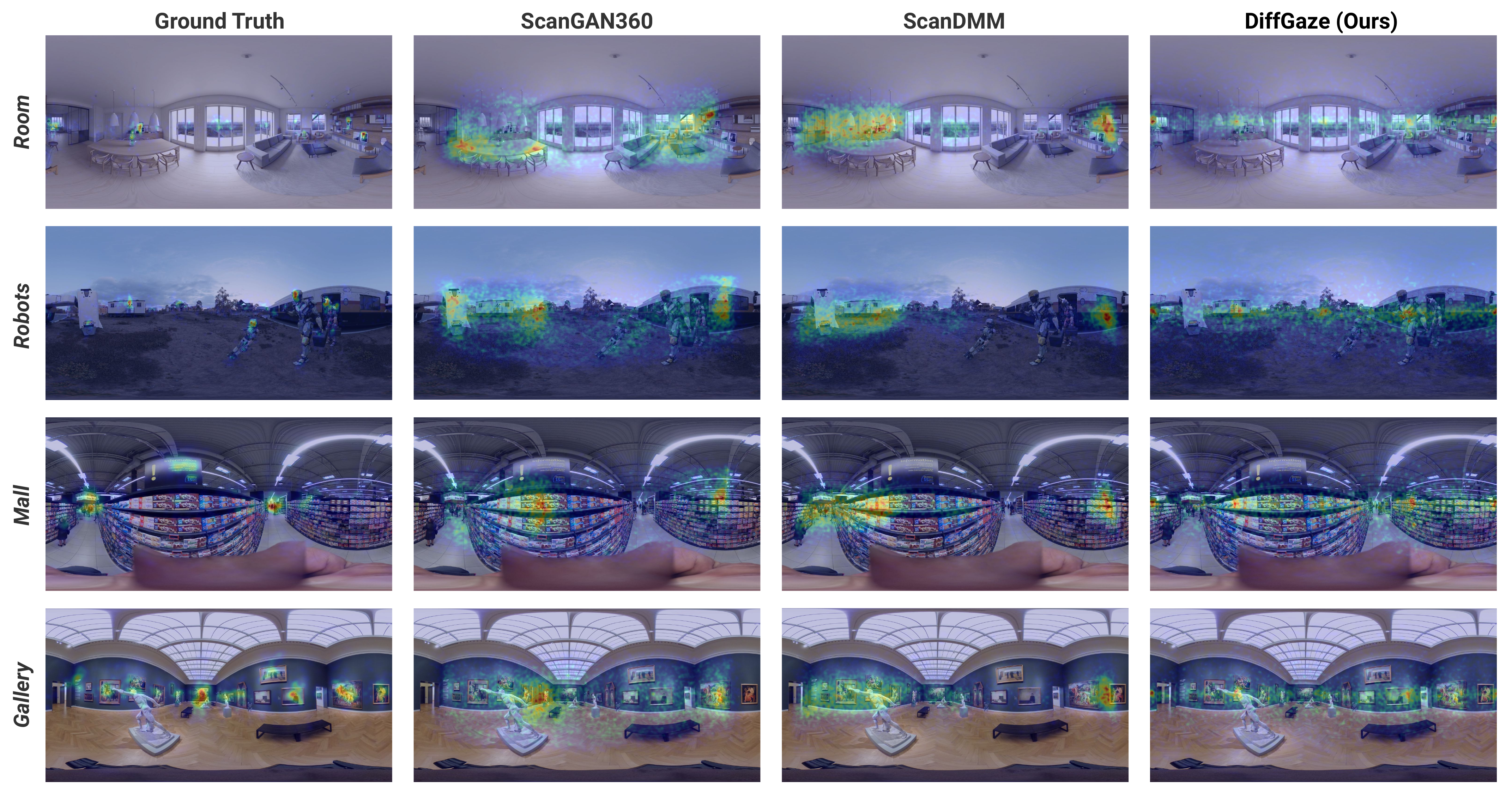}
    \caption{Qualitative comparison to saliency prediction models in four scenes. From left to right:  scanpaths obtained by a human observer, generated saliency maps obtained by ScanGAN360, ScanDMM, and the proposed model. From top to bottom: the Room and the Robots from Sitzmann dataset, the Mall and Gallery from Salient360! dataset.}
    \label{fig:experiment_saliency}
\end{figure*}

\textbf{Qualitative Results.}
\autoref{fig:experiment_saliency} shows sample saliency maps obtained from generated 30\,Hz gaze sequences of \methodName, ScanGAN360, and ScanDMM. See supplementary material for more examples. 
For instance, in the \textit{Gallery} scene, our method assigns high saliency to the sculpture and most of the paintings, while the other methods mainly focus on the sculpture and the rightmost painting. Similarly, in the \textit{Robots} scene, our method correctly identifies the salient regions, such as the largest robot and the robot in front of the cabin, while the other methods either miss them or include irrelevant regions, such as the whole cabin. These results demonstrate that our method can better capture the spatial location of human attention than the baselines.

\begin{table*}[t]
    \caption{Evaluation of saliency methods on Sitzmann and Salient360! dataset. The best results are shown in \textbf{bold}. } %
    \centering
    \begin{tabular}{lllllll}
\toprule
    \textbf{Dataset} & \textbf{Method} & \textbf{NSS}~$\uparrow$ & \textbf{CC}~$\uparrow$ & \textbf{AUC}~$\uparrow$ & \textbf{SIM}~$\uparrow$ & \textbf{KL}~$\downarrow$ \\ 
\midrule
\multirow{5}{*}{Sitzmann \cite{sitzmann2018saliency}}
    & BMS360~\cite{lebreton2018gbvs360} & \textbf{1.175}  & 0.509 & \textbf{0.883} & 0.421 & 0.225 \\ 
    & GBVS360~\cite{lebreton2018gbvs360} & 1.144 & \textbf{0.567} & 0.597 & \textbf{0.503} & \textbf{0.057} \\
    \cmidrule{2-7}
    & ScanDMM~\cite{sui2023scandmm} & 0.662  & 0.253 & 0.548 & 0.331 & 0.346 \\
    & ScanGAN360~\cite{martin2022scangan360} &  0.810 & 0.272 & 0.419 & 0.348 & 0.343 \\
    & \methodName (Ours) & 0.903 & 0.407	 & 0.812 & 0.358 & 0.318 \\
    \midrule
    
    \multirow{5}{*}{Salient360! \cite{salient360rai2017dataset, salient360rai2017saliency}} & BMS360~\cite{lebreton2018gbvs360} & \textbf{0.243} & \textbf{0.570} & \textbf{0.737} & \textbf{0.608} & 2.146 \\ 
    & GBVS360~\cite{lebreton2018gbvs360} & 0.201 & 0.562 & 0.601 & 0.583 & \textbf{1.609} \\ 
    \cmidrule{2-7}
    & ScanDMM~\cite{sui2023scandmm} & 0.149 & 0.368 & 0.454 & 0.476 & 2.468 \\
    & ScanGAN360~\cite{martin2022scangan360} &  0.119 & 0.260 & 0.411 & 0.400 & 3.337 \\
    & \methodName (Ours) & 0.176 & 0.468	 & 0.624 & 0.555 & 2.835 \\
\bottomrule
    \end{tabular}
    \label{table:sitzmann_saliency}
\end{table*}

\section{Discussion}

\subsection{Continuous Gaze Sequence \textit{vs.} Discrete Scanpath}

Previous works on scanpath prediction have focused on generating discrete sequences of fixations on 360$^\circ$ images. 
In contrast, we propose the first method to model and generate continuous gaze sequences that resemble the raw eye tracking data captured by real eye trackers.
Scanpaths only contain gaze fixation information while continuous gaze sequence includes both fixation and saccade signals.
Therefore, continuous gaze sequence generation can benefit not only the applications that rely on scanpath prediction, such as saliency prediction \cite{sui2023scandmm, martin2022scangan360} and image quality assessment \cite{sui2023scandmm}, but also many other applications. For example, computer graphics researchers have long sought to generate continuous gaze sequences for digital human animation. Existing methods rely on other inputs like head movements \cite{hu2019sgaze,hu2020dgaze} or speech \cite{Le2009speech}. Our method, \methodName, enables high frame rate animation of natural eye movements on a virtual human in a given scene. Compared to scanpath prediction approaches, it mimics more fine-grained human exploration of the image.
Another potential application of \methodName is to generate large-scale VR eye tracking datasets, which are difficult to collect due to the labour-intensive and privacy-compromising data collection process. Such datasets can help analyse human gaze behaviour in immersive environments and facilitate various gaze-based downstream tasks, such as eye event detection \cite{zemblys2019gazenet} or activity recognition \cite{Lan22EyeSyn}.

\subsection{Diffusion Model \textit{vs.} Other Methods}
In this work, we introduced a diffusion-based method for modelling continuous human gaze sequences.
We used two leading scanpath prediction methods -- a GAN-based approach \cite{martin2022scangan360} and a Deep Markov Model (DMM) \cite{sui2023scandmm} -- as strong baselines.
As demonstrated in \autoref{fig:experiment_rawgaze} and the supplementary material, both the GAN-based and DMM-based methods cannot generate plausible continuous human gaze data. 
This is due to the significantly higher complexity of continuous gaze data than scanpaths consisting of discrete fixation events.
Scanpath prediction models only need to account for fixation locations.
However, continuous gaze sequence generation models must account for various human eye movement events, including fixations and saccades. 
Moreover, generating a fixation involves predicting all raw gaze samples that form the fixation, not just a single location.
This requires a model to generate numerous gaze samples with minimal location shifts to form a fixation, then perform several saccades to the next fixation after a human-like duration, and repeat this process.

Different individuals may exhibit different eye movements on the same image, which complicates the training of an effective discriminator in ScanGAN360. 
Furthermore, ScanGAN360 incorporates the spherical DTW with the loss function of the GAN generator, which directly penalises the difference between the generated samples and one ground truth. Given that one image can have multiple, vastly different ground truths, ScanGAN360 exhibits an unusual pattern in modelling continuous eye movement data.
DMM models predict the likelihood of the next gaze location based on the current state information. As most gaze samples in continuous gaze sequences belong to fixations, these samples’ pattern has a higher likelihood in the prediction. Since the displacements of these samples are rare, ScanDMM tends to generate large clusters as a result (see \autoref{fig:experiment_rawgaze}). 

Our diffusion-based approach overcomes these limitations by modelling the complex distribution of continuous gaze data through a forward and backward process. 
Unlike ScanGAN360, our method does not require a discriminator or a distance metric to train the generator, which makes it easier to train and more robust to different ground truths.
Unlike ScanDMM, our method generates the whole gaze sequence at once, which enables it to capture the global spatial and temporal coherence of human gaze behaviour. 
We demonstrate that our method can generate continuous gaze sequences that match the statistics and characteristics of human gaze data and are visually indistinguishable from real human gaze sequences. 

\subsection{Performance Metrics}

Our work also sheds light on the challenge of evaluating the performance of attention models, as well as the shortcomings of widely used evaluation metrics and their discrepancies to human visual judgement.
As discussed in \autoref{sec:gaze_seq_exp}, human agreement performs poorly on selected time-series metrics due to the significant differences between real human gaze sequences. Additionally, we observed disagreements between the Levensthein distance and qualitative evaluation results. Specifically, ScanDMM performed the best in the best LEV (see \autoref{table:raw_quan}) but the generated gaze samples are visually implausible and received very low user ratings for realism in our user study. 
Interestingly, we also observed the disagreement between the qualitative results of generated gaze sequences and scanpath metrics. As shown in \autoref{table:scanpath_quan}, the extracted scanpaths of ScanDMM and ScanGAN360 achieve good overall performance. Especially, for ScanGAN360, the generated 30 Hz gaze sequences are entirely visually unrealistic according to our user study result. This indicates that evaluating the scanpaths extracted from generated gaze sequences cannot reflect the method’s performance on continuous gaze sequence generation.
Statistics at the eye movement level (\autoref{table:raw_stat}) and the user study results are in line with our visual intuitions. However, the spatial and temporal aspects are not shown in statistics. Besides, having a user study to evaluate a large scale of generated continuous trajectories is highly impractical. Therefore, we see an urgent need for designing effective evaluation metrics that we plan to explore in future work.

\subsection{Limitations and Future Work}
The Sitzmann~\cite{sitzmann2018saliency} and Salient360!~\cite{salient360rai2017dataset, salient360rai2017saliency} datasets used in our work only contain 107 360$^\circ$ images, and we used 19 images to train \methodName. 
The limited training data size may make the generalisability of our model questionable. 
However, these two the only public 360$^\circ$ datasets providing raw gaze data. \methodName can also be modified to generate continuous gaze sequences for other tasks, e.g. visual search \cite{chen2021coco, Yang_2020_CVPR}, visual question answering \cite{chen2020air}, and natural images \cite{Judd_2009, borji2015cat2000}. However, among these, only the MIT1003 dataset \cite{Judd_2009} provides raw gaze data.
More datasets with raw eye tracking data would enhance models like \methodName. We encourage the research community to make such data public.

In addition, running time-series metrics on high-frequency gaze data is time-intensive, we only tested our method in generating 30\,Hz gaze trajectories. Future work will explore our model’s performance on higher sampling frequencies to see whether it can model other eye events. To detect certain types of eye events, such as microsaccades, prior work \cite{nystrom2021tobii} suggests using the highest available sampling frequency, such as 1,000\,Hz, to capture the eye tracking data. However, current 360$^\circ$ image eye tracking datasets typically have a maximum sampling frequency of 120\,Hz. We anticipate the availability of more high-frequency VR eye tracking datasets to enhance continuous gaze sequence generation.

\section{Conclusion}

This paper introduced \methodName, a conditional diffusion model for generating realistic and diverse continuous human gaze sequences in 360$^{\circ}$ environments. This method significantly advances the field by moving beyond the prediction of scanpaths to model more complex eye movements. The effectiveness of \methodName was demonstrated through rigorous evaluation on two 360$^{\circ}$ image datasets across three different tasks. Not only did \methodName outperform previous methods in terms of gaze sequence generation, scanpath prediction and saliency prediction, but it also showed comparable performance with human baseline, underscoring its ability to simulate human-like gaze behaviour. These results highlight the potential of \methodName to facilitate further research in gaze behaviour analysis in immersive environments. By providing high-quality simulated eye-tracking data, \methodName opens up new possibilities for human-computer interaction and computer vision applications, paving the way for more intuitive and immersive user experiences.

\ifCLASSOPTIONcompsoc
  \section*{Acknowledgments}
\else
  \section*{Acknowledgment}
\fi

C. Jiao was funded by the EU's Horizon Europe research and innovation funding programme under grant agreement No. 101072410 (Eyes4ICU). 
Y. Wang was funded by the Deutsche Forschungsgemeinschaft (DFG, German Research Foundation) -- Project-ID 251654672 -- TRR 161. 
M. Bâce was funded by the Swiss National Science Foundation (SNSF) through a Postdoc.Mobility Fellowship (grant number 214434) while at the University of Stuttgart.
Z. Hu was funded by the Deutsche Forschungsgemeinschaft (DFG, German Research Foundation) under Germany's Excellence Strategy -- EXC 2075 -- 390740016.
A. Bulling was funded by the European Research Council (ERC) under grant agreement 801708.

\ifCLASSOPTIONcaptionsoff
  \newpage
\fi

We would like to thank 
the International Max Planck Research School for Intelligent Systems (IMPRS-IS) for supporting G. Zhang.

\bibliographystyle{IEEEtranN}
\bibliography{main}

\section{Biography Section}
\vspace{-25pt}

\begin{IEEEbiography}[{\includegraphics[width=1in,height=1.25in,clip,keepaspectratio]{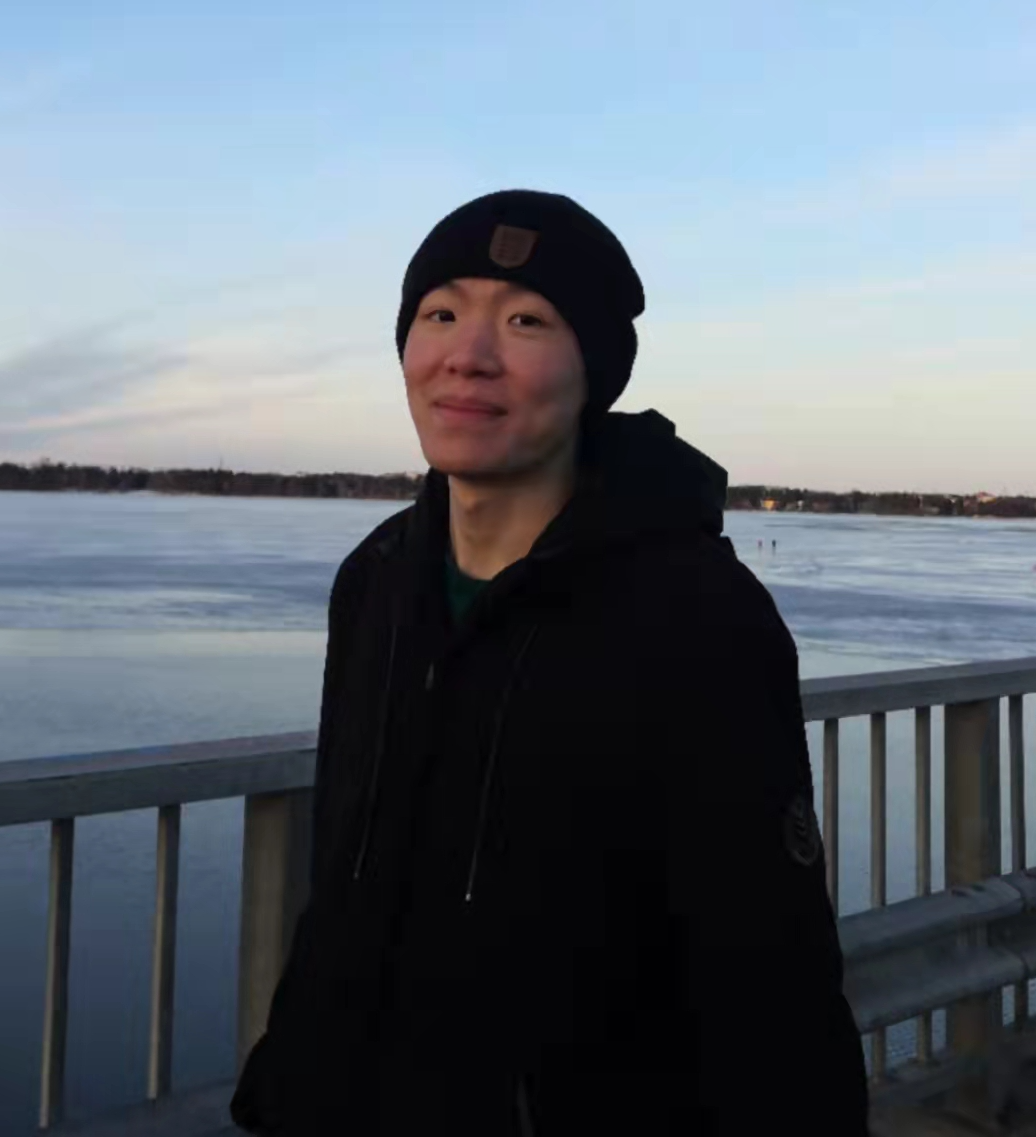}}]{Chuhan Jiao}
is a PhD student at the University of Stuttgart, Germany. He received his BEng. in Computer Science and Technology from Donghua University, China, in 2020 and MSc. in Computer Science from Aalto University, Finland, in 2022. His research interest lies at the intersection of computer vision and human-computer interaction. 
\end{IEEEbiography}

\vspace{-30pt}

\begin{IEEEbiography}[{\includegraphics[width=1in,height=1.25in,clip,keepaspectratio]{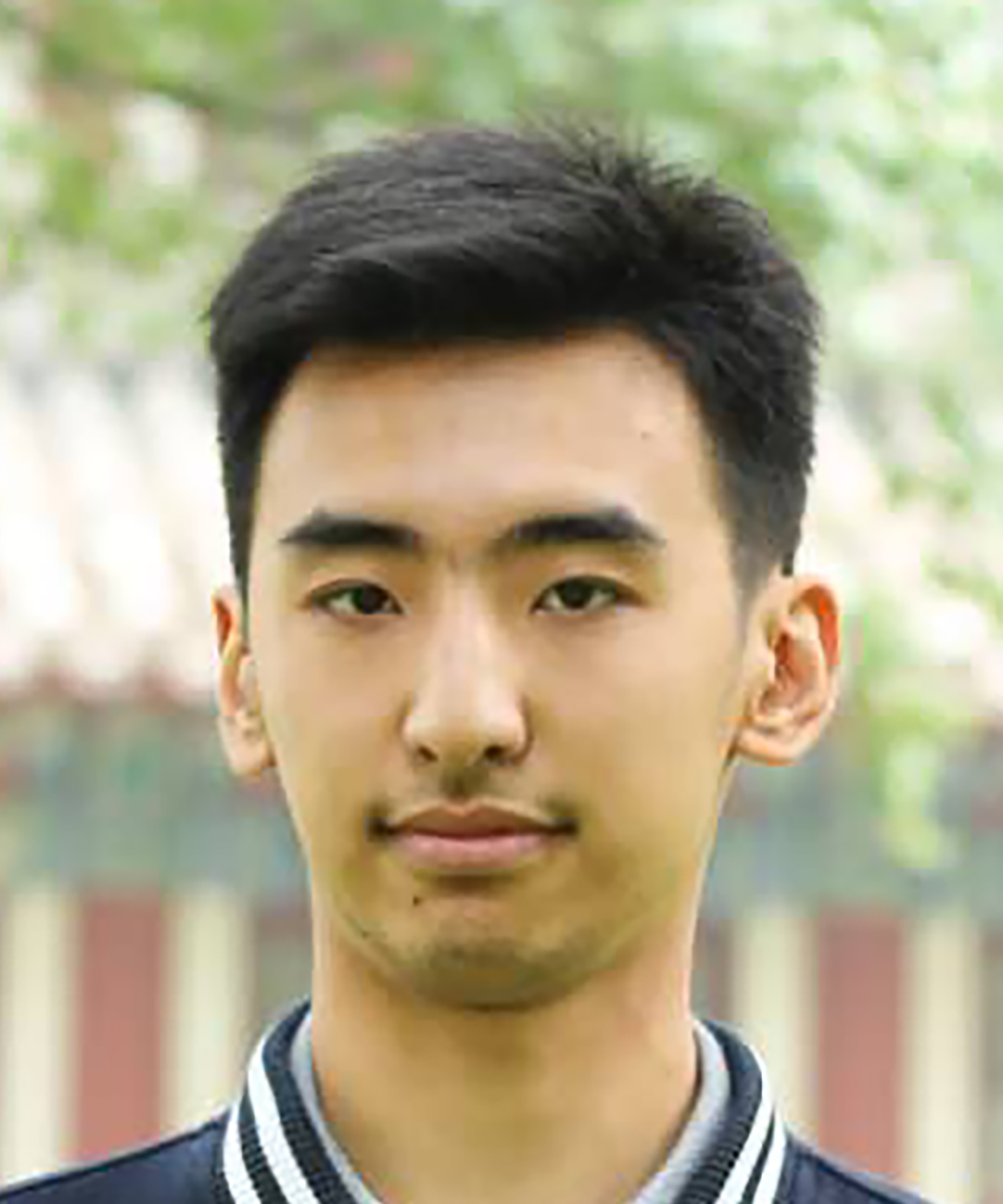}}]{Yao Wang}
is a PhD student at the University of Stuttgart, Germany. He received his BSc. in Intelligence Science and Technology and MSc. in Computer Software and Theory both from Peking University, China, in 2017 and 2020, respectively. His research interests include computer vision and human-computer interaction, with a focus on visual attention modelling on information visualizations.
\end{IEEEbiography}

\vspace{-25pt}

\begin{IEEEbiography}[{\includegraphics[width=1in,height=1.25in,clip,keepaspectratio]{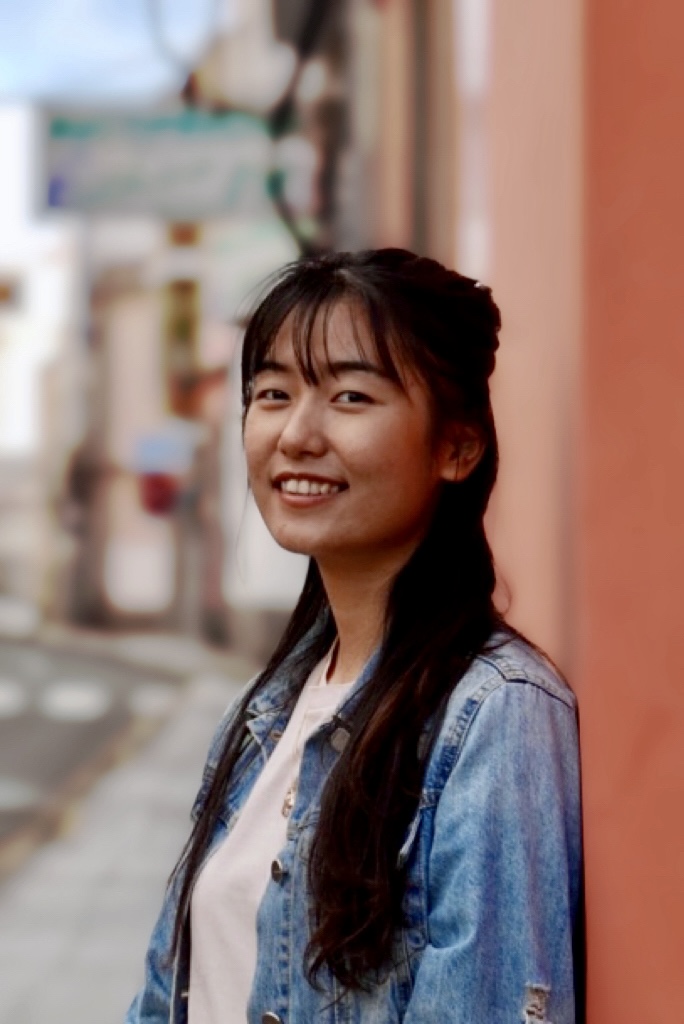}}]{Guanhua Zhang}
is a PhD student at the University of Stuttgart, Germany. She received her Master's degree in Computer Science and Technology from Tsinghua University in 2020 and Bachelor's degree in Computer Science and Technology from Beijing University of Posts and Telecommunications in 2017. Her research interests include computational interaction, affective computing, and representation learning.
\end{IEEEbiography}

\vspace{-25pt}

\begin{IEEEbiography}[{\includegraphics[width=1in,height=1.25in,clip,keepaspectratio]{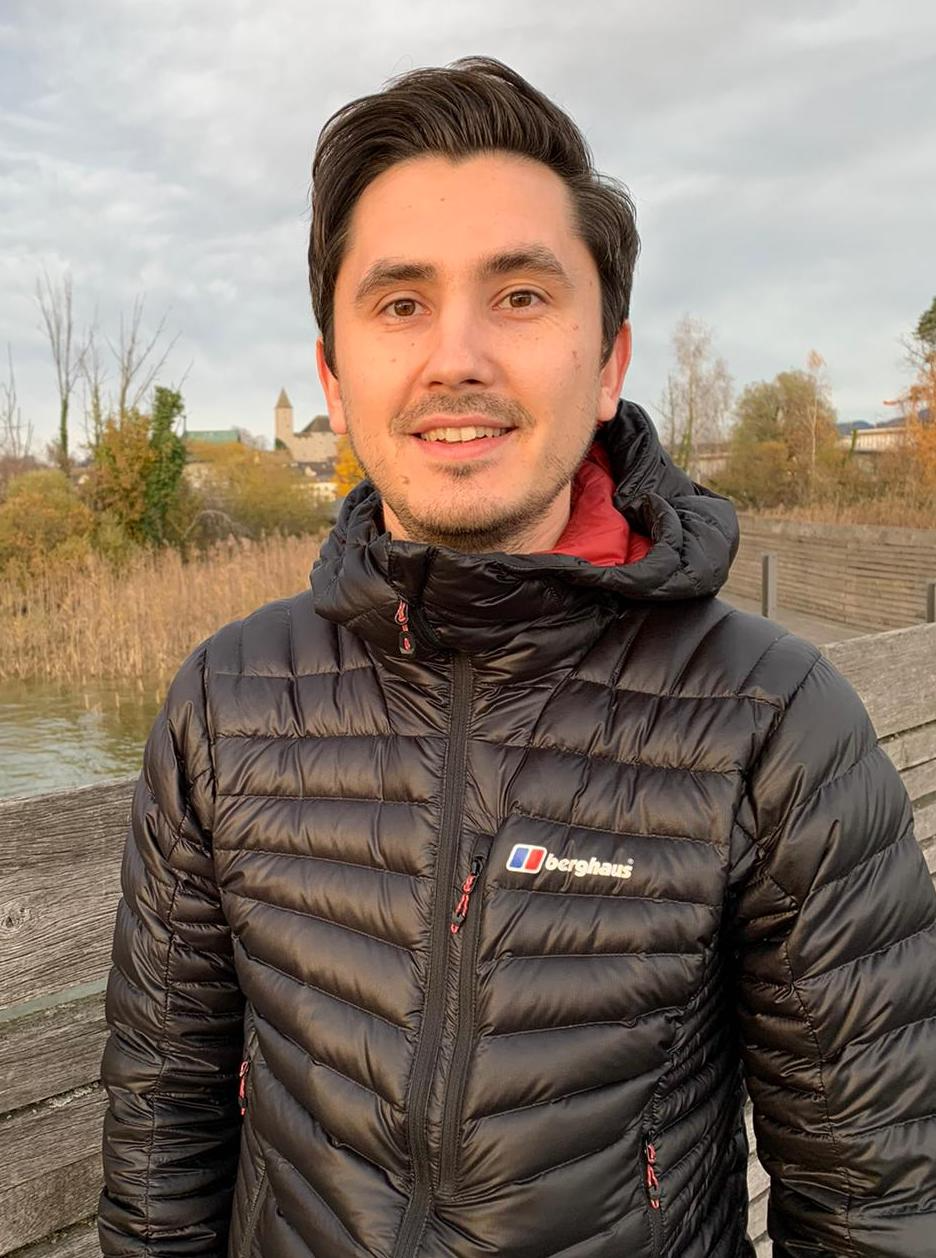}}]{Mihai Bâce}
is an assistant professor in the Department of Computer Science and member of the \mbox{e-Media} Research Lab at KU Leuven, Belgium. He received a PhD in Computer Science from ETH Zurich, Switzerland in 2020, a MSc. in Computer Science from EPFL, Switzerland, in 2014 and his BSc. in Computer Science from the Technical University of Cluj-Napoca, Romania in 2012. His research interests are at the intersection of machine learning and human-computer interaction with a focus on computational user modelling.
\end{IEEEbiography}

\vspace{-25pt}

\begin{IEEEbiography}[{\includegraphics[width=1in,height=1.25in,clip,keepaspectratio]{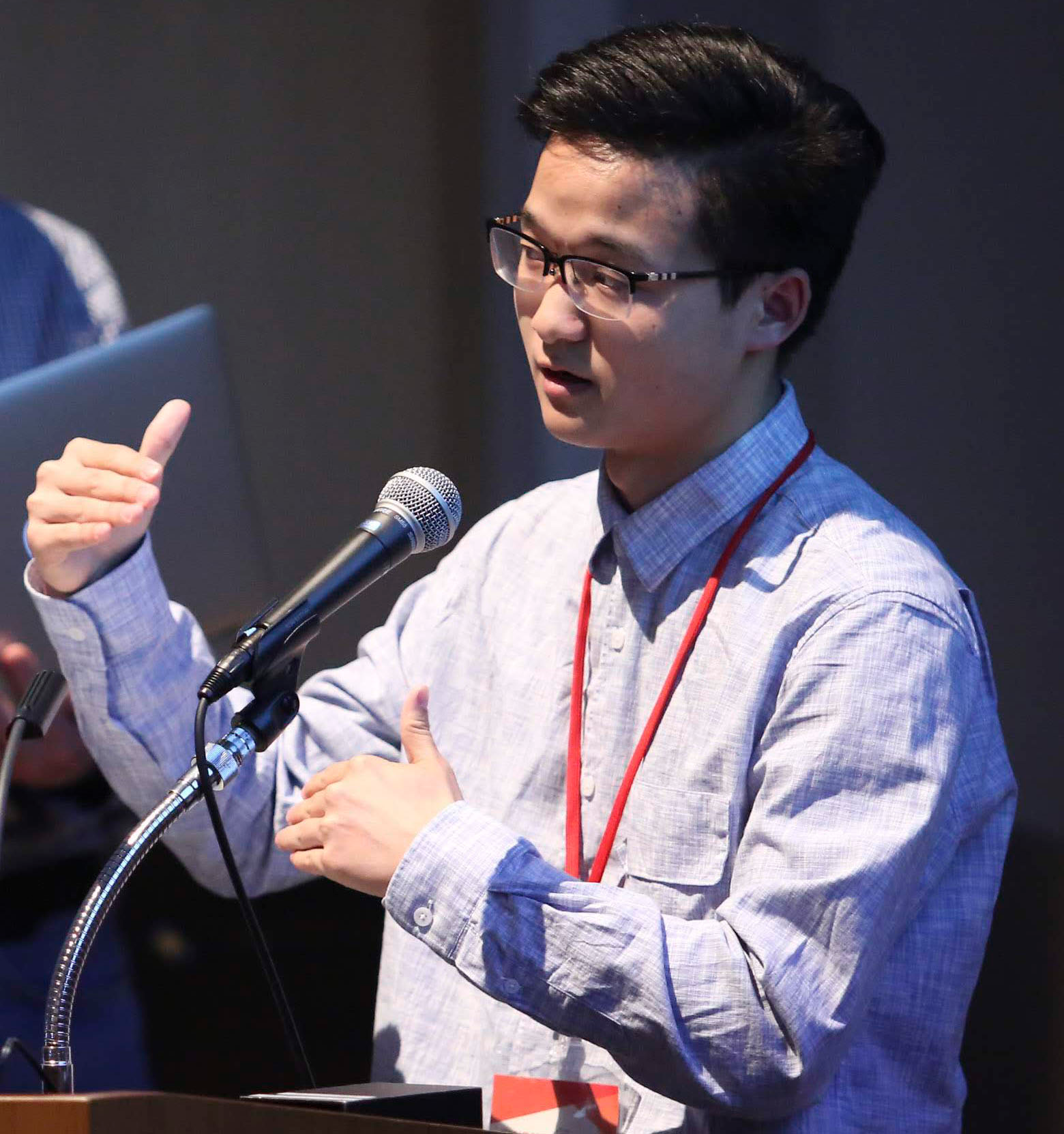}}]{Zhiming Hu} 
is a post-doctoral researcher at the University of Stuttgart, Germany. 
He obtained his Ph.D. degree in Computer Software and Theory from Peking University, China in 2022 and received his Bachelor's degree in Optical Engineering from Beijing Institute of Technology, China in 2017. His research interests include virtual reality, human-computer interaction, eye tracking, and human-centred artificial intelligence.
\end{IEEEbiography}

\vspace{-30pt}

\begin{IEEEbiography}[{\includegraphics[width=1in,height=1.25in,clip,keepaspectratio]{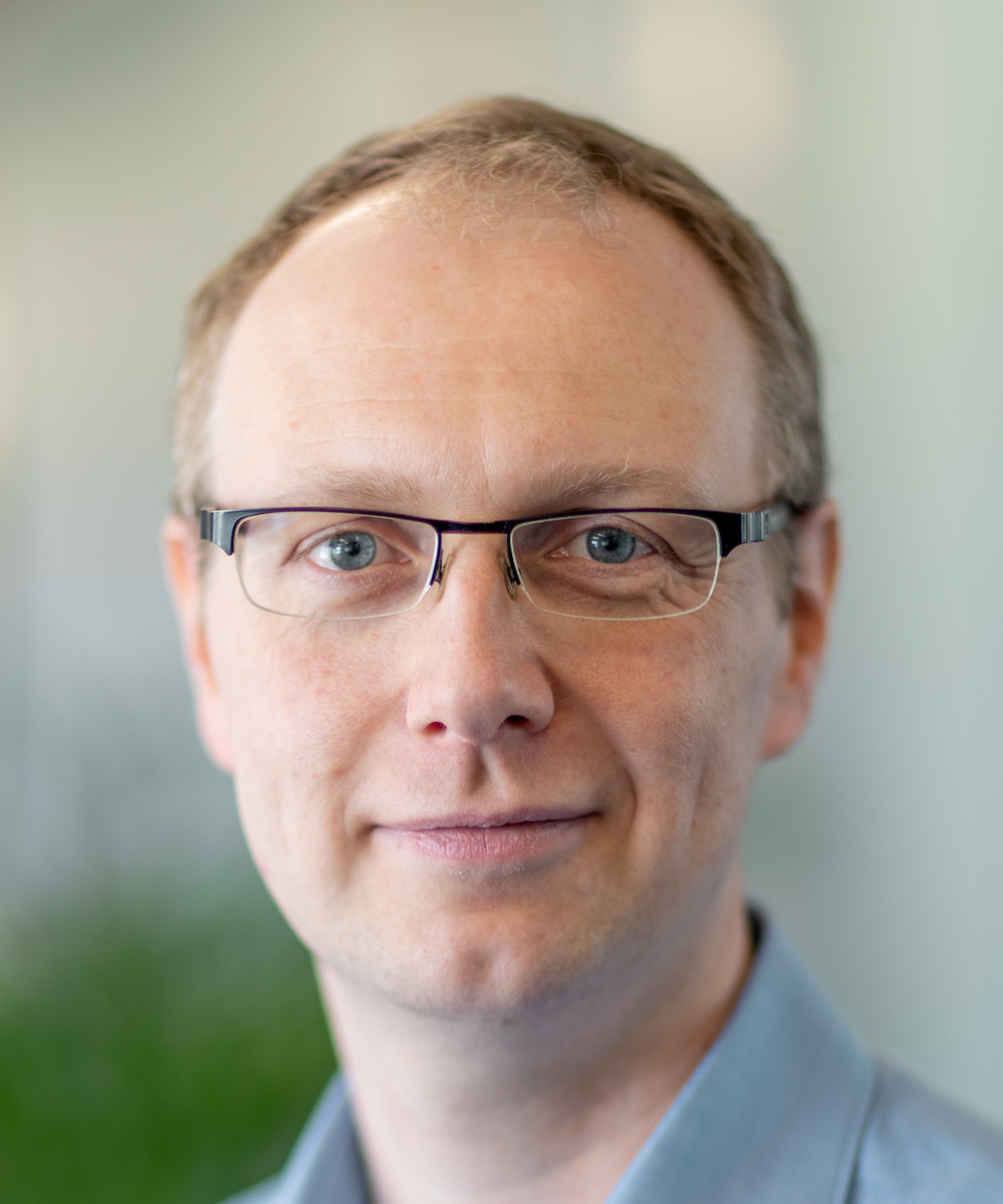}}]{Andreas~Bulling}
is Full Professor of Computer Science at the University of Stuttgart, Germany, where he directs the research group "Human-Computer Interaction and Cognitive Systems". He received his MSc. in Computer Science from the Karlsruhe Institute of Technology, Germany, in 2006 and his PhD in Information Technology and Electrical Engineering from ETH Zurich, Switzerland, in 2010. Before, Andreas Bulling was a Feodor Lynen and Marie Curie Research Fellow at the University of Cambridge, UK, and a Senior Researcher at the Max Planck Institute for Informatics, Germany. His research interests include computer vision, machine learning, and human-computer interaction.
\end{IEEEbiography}

\vfill

\end{document}